\documentclass[10pt,twocolumn,letterpaper]{article}

\usepackage{iccv}
\usepackage{times}
\usepackage{epsfig}
\usepackage{graphicx}
\usepackage{amsmath}
\usepackage{amssymb}
\usepackage{booktabs}
\usepackage{multirow}
\usepackage[table]{xcolor}
\definecolor{grayrow}{cmyk}{0,0,0,0.18}

\usepackage[breaklinks=true,bookmarks=false,colorlinks]{hyperref}

\iccvfinalcopy 

\def\iccvPaperID{9766} 

\ificcvfinal\fi

\begin{document}

\title{
GLA-GCN: Global-local Adaptive Graph Convolutional Network for 3D Human Pose Estimation from Monocular Video
}

\author{Bruce X.B. Yu\textsuperscript{1}\qquad
Zhi Zhang\textsuperscript{1}\qquad
Yongxu Liu\textsuperscript{1}\quad
\and
Sheng-hua Zhong\textsuperscript{2}\qquad
Yan Liu\textsuperscript{1}\qquad
Chang Wen Chen\textsuperscript{1}\\ \\
\textsuperscript{1}The Hong Kong Polytechnic University\qquad
\textsuperscript{2}Shenzhen University
}

\maketitle
\ificcvfinal\thispagestyle{empty}\fi

\begin{abstract}
3D human pose estimation has been researched for decades with promising fruits. 3D human pose lifting is one of the promising research directions toward the task where both estimated pose and ground truth pose data are used for training. Existing pose lifting works mainly focus on improving the performance of estimated pose, but they usually underperform when testing on the ground truth pose data. We observe that the performance of the estimated pose can be easily improved by preparing good quality 2D pose, such as fine-tuning the 2D pose or using advanced 2D pose detectors. As such, we concentrate on improving the 3D human pose lifting via ground truth data for the future improvement of more quality estimated pose data.
Towards this goal, a simple yet effective model called Global-local Adaptive Graph Convolutional Network (GLA-GCN) is proposed in this work. Our GLA-GCN globally models the spatiotemporal structure via a graph representation and backtraces local joint features for 3D human pose estimation via individually connected layers.
To validate our model design, we conduct extensive experiments on three benchmark datasets: Human3.6M, HumanEva-I, and MPI-INF-3DHP. Experimental results show that our GLA-GCN\footnote{Code is available: \url{https://github.com/bruceyo/GLA-GCN} } implemented with ground truth 2D poses significantly outperforms state-of-the-art methods (e.g., up to 3\%, 17\%, and 14\% error reductions on Human3.6M, HumanEva-I, and MPI-INF-3DHP, respectively). 
\end{abstract}

\section{Introduction}
3D Human Pose Estimation (HPE) in videos aims to predict the pose joint locations of the human body in 3D space, which can facilitate plenty of applications such as video surveillance, human-robot interaction, and physiotherapy \cite{sarafianos20163d}. 3D human poses can be directly retrieved from advanced motion sensors such as motion capture systems, depth sensors, or stereotype cameras \cite{RN001,RN002}. The 3D HPE task can be performed under either multi-view or monocular view settings. Although state-of-the-art multi-view methods \cite{iskakov2019learnable, reddy2021tessetrack, zhang2021adafuse,he2020epipolar} generally show superior performance than monocular ones \cite{hu2021conditional,zhang2022mixste}, ordinary RGB monocular cameras are much cheaper than these off-the-shelf motion sensors and more widely applied in real-world surveillance scenarios. Hence, 3D HPE from a monocular video is an important and challenging task, which has been attracting increasing research interest. 
Recent monocular view works can be grouped into model-based and model-free methods \cite{RN007}. Model-based methods \cite{cheng20203d, gong2021poseaug} incorporate parametric body models such as kinematic \cite{RN015}, planar \cite{RN016}, and volumetric models \cite{RN017} for 3D HPE. Model-free methods can be further grouped into single-stage and 2D to 3D lifting methods. Single-stage methods estimate the 3D pose directly from images in an end-to-end manner \cite{RN008,RN010,chen2021deductive, wehrbein2021probabilistic, ma2021context, RN021}. 2D to 3D lifting methods have an intermediate 2D pose estimation layer \cite{RN011,RN012,RN013,RN014}.
Among these methods, 2D to 3D lifting methods implemented with ground truth 2D poses achieved better performance.

The advantages of 2D to 3D lifting methods can be summarized as two main points: allowing make use of advances in 2D human pose detection and exploiting temporal information along multiple 2D pose frames \cite{RN012,ji2020survey}. For the 2D human pose detection, it has achieved remarkable progress via detectors such as Mask R-CNN (MRCNN) \cite{RN004}, Cascaded Pyramid Network (CPN) \cite{RN005}, Stacked Hourglass (SH) detector \cite{RN003},  and HR-Net \cite{sun2019deep}. The intermediate 2D pose estimation stage via these 2D pose detectors significantly reduces the data volume and complexity of the 3D HPE task. For the temporal information, existing mainstream methods \cite{RN012,RN013,RN014, hu2021conditional, li2022exploiting, zhang2022mixste} gained noticeable improvements by feeding a long sequence of 2D pose frames to their models, among which \cite{zhang2022mixste} achieved the state-of-the-art performance via ground truth 2D poses. Recent methods \cite{zhang2022mixste, motionbert2022} simply fine-tuned these 2D pose detectors on the target datasets and achieved great improvements for the performance of estimated 2D pose data but remain far behind the results of using ground truth 2D pose, which motivates us to concentrate on improving the 3D HPE via ground truth 2D pose data for potential improvements via future more quality estimated 2D pose data.  

Given the promising performance and advantages of 2D to 3D lifting methods, our work contributes the literature along this direction. For 2D to 3D lifting approaches, since \cite{RN011} proposed Fully Connected Network (FCN), recent advanced models have three main groups: Temporal Convolutional Network (TCN)-based \cite{RN012,RN013},  Graph Convolutional Network (GCN)-based \cite{RN018,RN014, hu2021conditional}, and Transformer-based ones \cite{li2022mhformer, li2022exploiting, zhang2022mixste}. 
On the one hand, we observe that existing TCN- and Transformer-based methods can receive large receptive fields (i.e., a long 2D pose sequence) with strided convolutions. However, it can be difficult to make further intuitive designs to backtrace local joint features based on the pose structure, since the 2D pose sequence is flattened and fed to the model. Meanwhile, the estimation of different pose joints relies on the same fully connected layer, which lacks considering the independent characteristic of different pose joints. On the other hand, GCN-based models can explicitly reserve the structure of 2D and 3D human pose during convolutional propagation. However, this advantage of GCN remains under-explored. Existing GCN-based methods \cite{RN018,RN014} also utilized a fully connected layer for the estimation of different 3D pose joints, which does not consider the structural features of GCN representations. 

To this end, we propose Global-local Adaptive GCN (GLA-GCN) for 2D to 3D human pose lifting. Our GLA-GCN contains two modules: global representation and local 3D pose estimation. 
In the global representation, we use an adaptive Graph Convolutional Network (GCN) to reconstruct the global representation of an intermediate 3D human pose sequence from its corresponding 2D sequence. For the local 3D pose joint estimation, we temporally shrink the global representation optimized by the reconstructed 3D pose sequence with a strided design. Then, an individual connected layer is proposed to locally estimate the 3D human pose joints from the shrunken global representation. 
Our contributions can be threefold as follows:

$\bullet$ We propose a global-local learning architecture that leverages the global spatialtemporal representation and local joint representation in the GCN-based model for 3D human pose estimation.

$\bullet$ We are the first to introduce an individual connected layer that has two components to divide joint nodes and input the joint node representation for 3D pose joint estimation instead of based on pooled features.

$\bullet$ Our GLA-GCN model performs better than corresponding state-of-the-art methods \cite{li2022exploiting,zhang2022mixste} with considerable margins e.g., up to 3\% and 17\% error reductions on Human3.6M \cite{RN025} and HumanEva-I \cite{RN033}, respectively.

\section{Related Work}
\noindent \textbf{2D to 3D Lifting}.
3D HPE is a traditional vision problem that has been studied for decades \cite{dang2019deep,zeng2021learning,cai2019exploiting,wang2019generalizing,gartner2022trajectory,kundu2022uncertainty,wang2022distribution,wei2022capturing,wang2022ocr,liu2021multi}. Existing works of 3D HPE from a monocular view usually target two main scenarios: single person and multi-person \cite{RN007}. This work aims to improve the performance of single person 3D HPE. \cite{RN022,RN023,RN024} represent early efforts that attempt to infer 3D position from 2D projections. They usually rely on manually chosen parameters based on assumptions about pose joint mobility. Methods \cite{gong2021poseaug,zhan2022ray3d} estimating 3D pose from less frames or even a single frame has shown great progress but can be a lack of considering temporal information. Recent advances in 2D human pose estimation \cite{RN003,RN004,RN005} enable 2D to 3D lifting approaches to achieve remarkable performance over other counterparts.
Inspired by \cite{RN011}, there has been more well-designed learning architectures being proposed to improve the performance, in particular, by utilizing temporal information. These methods are also known as 2D to 3D lifting, which can be grouped into three directions: TCN-, GCN-, and Transformer-based architectures \cite{RN012,RN013,RN018,RN014, hu2021conditional,li2022exploiting, zhang2022mixste}.

TCN-based methods \cite{RN012, RN013} successfully push the performance of 2D to 3D lifting forward with a strided design for their learning architectures built upon 1D CNN layers. The strided design is on the temporal dimension of the input, which allows the features to shrink from a 2D pose sequence to a feature embedding for the 3D pose estimation via a final fully connected layer. The number of channels for the fully connected layer is conventionally set to $1024$, which is shared to predict the 3D positions of all pose joints. While varied numbers of input 2D pose frames have been extensively investigated, which shows input 2D pose frames with reasonable length can benefit the 3D pose reconstruction. The strided design can effectively reduce the feature size by shrinking the number of temporal frames along propagation of several TCN blocks. Using this strided structure, Transformer-based methods \cite{li2022exploiting, zhang2022mixste} show promising performance, especially \cite{zhang2022mixste} that takes advantage of weighted and temporal loss functions and helps it outperform the GCN-based methods optimized with an additional motion loss \cite{RN014, hu2021conditional}. The motion loss was shown not very effective in \cite{zhang2022mixste}. These observations compel us to explore effective models in the direction of GCN-based models with the inspiring designs in mind but without relying on various novel loss functions.

\begin{figure*}
	\begin{center}
		\includegraphics[width=0.84\linewidth]{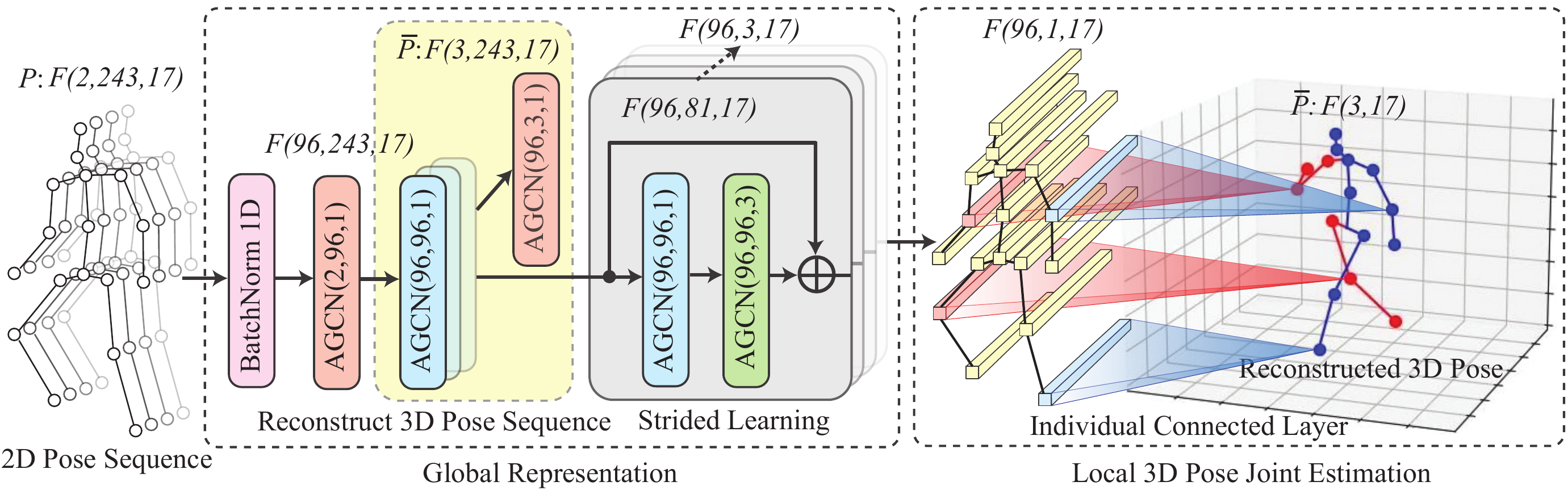}
	\end{center}
 \vspace{-8pt}
	\caption{Learning architecture of our GLA-GCN. $AGCN(C_{in},C_{out},S)$ represents AGCN blocks with the specific values of the input channel, output channel, and stride length. $F(C',T',N')$ represents the size of a feature map. The individual connected layer shows the prediction process of four pose joint examples that use separate 1D CNN layers. 
	}
	\label{fig:architecture_SAGCN}
  \vspace{-5pt}
\end{figure*}

\noindent \textbf{Graph Convolutional Network}.
A popular method representing the pose data with GCN is Spatial Temporal GCN (ST-GCN) \cite{RN026}, which is originally proposed to model large receptive fields for the skeleton-based action recognition. Following ST-GCN, advanced GCN models have been proposed to advance 3D HPE \cite{RN029, RN018, RN020,RN014}. 

Regarding GCN-based models for 3D HPE, Ci et al.  \cite{RN029} proposed Locally Connected Network (LCN) that takes the advantages of FCN \cite{RN011} and GCN \cite{RN030}. LCN has the similar design for the convolutional filters to ST-GCN \cite{RN026}, which defines a neighbor set for a node based on the distance to perform convolutional operation. Zhao et al. \cite{RN018} proposed an architecture called SemGCN that stacks GCN layers by flatten output to a fully connected layer. The optimization of SemGCN is based on both joint positions and bone vectors. Choi et al. \cite{RN020} also proposed to use GCN to recover 3D human pose and mesh from a 2D human pose. Liu et al. \cite{RN031} investigated how weight sharing schemes in GCNs affect the pose lifting task, which shows the pre-aggregation method leads to relatively better performance. The architecture in \cite{RN031} is similar with that of SemGCN. The above mentioned GCN-based methods achieved good performance via a single pose frame input but they did not take the advantage of temporal information in a 2D pose sequence. 

Taking multiple 2D pose frames as input, U-shaped Graph Convolution Networks (UGCN) \cite{RN014, hu2021conditional} further improves the performance of GCN-based methods by paying attention to the temporal characteristics of a pose motion. Specifically, UGCN utilizes spatial temporal GCN \cite{RN026} to predict a 3D pose sequence from a 2D pose sequence for the reconstruction of a single 3D pose frame. A motion loss term that regulates the temporal trajectory of pose joints based on the prediction of a 3D pose sequence and its corresponding ground truth 3D pose sequence. Despite the improvements grained with novel loss terms in works such as SemGCN and UGCN, we aim to contribute the literature of 2D-3D lifting by using the consistent loss term used in \cite{RN012,RN013}. In our model design, we propose to incorporate the strided convolutions to a GCN-based model that represents global information of a 2D pose sequence. Based on the structure of GCN representation, we explicitly utilize the structured features of different pose joints to locally predict their corresponding 3D pose locations. 

\section{Method}
Given the temporal information of a 2D human pose sequence estimated from a video $P=\{p_{t,i}\in\mathbb{R}^{2}|\ t=1,...,T;i=1,...,N\}$, where $T$ is the number of pose frames and $N$ is the number of pose joints, we aim to utilize this 2D pose sequence $P$ to reconstruct 3D coordinates of pose joints $\bar{P}=\{\bar{p}_{i}\in\mathbb{R}^{3}|i=1,...,N\}$. Figure \ref{fig:architecture_SAGCN} shows the learning architecture of our GLA-GCN, which uses AGCN layers to globally represent the 2D pose sequence and locally estimate the 3D pose via an individual connected layer. In the following of this section, we introduce the detailed design of our GLA-GCN.
\subsection{Global Representation}
\noindent \textbf{Adaptive Graph Convolutional Network}.
An AGCN block \cite{li2018adaptive,RN027} is based on the GCN with an adaptive design that improves the flexibility of a typical ST-GCN block \cite{RN026}. 
Let us represent the 2D pose sequence $P$ as a spatial-temporal graph $\mathcal{G}=\{{\upsilon}_{t},\varepsilon_{t}|t=1,...,T\}$, where $\upsilon_{t}=\{\upsilon_{t,i}|i=1,...,N\}$ represents the pose joints and $\varepsilon_{t}$ represents the corresponding pose bones. To implement a basic ST-GCN block, a neighbor set $\mathcal{B}_{i}$ is first defined to indicate the spatial graph convolutional filter for a specific pose joint $\upsilon_{t,i}$. Specifically, for the graph convolutional filter of a vertex node, we apply three distance neighbor subsets: the vertex itself, centripetal subset, and centrifugal subset. The definitions of centripetal and centrifugal subsets are based on the pose frame’s gravity center (i.e., the average coordinate of all pose joints). Centripetal and centrifugal subsets represent nodes that are closer and farther to the average distance from the gravity center, respectively. Empirically, similar with 2D convolution, we set the kernel size $K$ to $3$, which will lead to $3$ subsets in $\mathcal{B}_{i}$. To implement the subsets, a mapping $h_{t,i}\rightarrow\{0,...,K-1\}$ is used to index each subset with a numeric label, where centripetal and centrifugal subsets are respectively labeled as $1$ and $2$. Subsets that have the average distance to gravity center is indexed to $0$. 
This graph convolutional operation can be written as
\begin{equation} \label{eq:1}
f_{out}({\upsilon}_{t,i})=\sum\nolimits_{\upsilon_{t,j} \in \mathcal{B}_{i}}\frac{1}{Z_{t,j}}f_{in}\left(\upsilon_{t,j}\right)W(h_{t,i}(\upsilon_{t,j}))
\end{equation}
where $f_{in}:v_{t,j}\rightarrow \mathbb{R}^2$ is a mapping that gets the attribute features of joint node $v_{t,j}$ and $Z_{t,j}$ is a normalization term that equals to the subset’s cardinality. $W(h_{t,i}(v_{t,j}))$ is a weight function $W\left(\upsilon_{t,i},\upsilon_{t,j}\right):\mathcal{B}_{i}\rightarrow \mathbb{R}^2$ implemented by indexing a $\left(2,K\right)$ tensor.
For a pose frame, the determined graph convolution of a sampling strategy (e.g., centripetal and centrifugal subsets) can be implemented by an $N\times N$ adjacency matrix. Specifically, with $K$ spatial sampling strategies $\sum_{k=1}^K\mathbf{A}_k$ and the adaptive design, Equation \ref{eq:1} can be transformed into
\begin{equation} \label{eq:2}
\mathbf{f}_{out}(\upsilon_{t})=\sum\nolimits_{k=1}^K (\mathbf{A}_k+\mathbf{B}_k+\mathbf{C}_k) \mathbf{f}_{in}\mathbf{W}_k 
\end{equation}
where $\mathbf{\Lambda}_k^{-\frac{1}{2}}\mathbf{\bar{A}}_k\mathbf{\Lambda}_k^{-\frac{1}{2}}$ is a normalized adjacency matrix of $\mathbf{\bar{A}}_k$ with its elements indicating whether a vertex $\upsilon_{t,j}$ is included in the neighbor set. $\mathbf{\Lambda}_k^{ii}=\sum_j(\mathbf{\bar{A}}_k^{ij})+\alpha$ is a diagonal matrix with $\alpha$ set to $0.001$ to prevent empty rows. $\mathbf{W}_k$ denotes the weighting function of Equation \ref{eq:1}, which is a weight tensor of the $1\times 1$ convolutional operation. Unlike $\mathbf{A}_k$ that represents the physical structure of a human pose, $\mathbf{B}_k$ represents learnable parameters that indicate the connection strength between pose joints,
which is implemented with an $N\times N$ adjacency matrix initialized to $0$. $\mathbf{C}_k$ performs the similar function of $\mathbf{B}_k$, which is implemented by the dot product of two feature maps calculated by embedding functions (i.e., $\theta$ and $\phi$) to calculate the similarity between pose joints. Calculation of $\mathbf{C}_k$ can be represented as 
\begin{equation} \label{eq:3}
\mathbf{C}_k\ =\ SoftMax(\mathbf{f}_{in}^T\mathbf{W}_{\theta k}^T\mathbf{W}_{\phi k}\mathbf{f}_{in})
\end{equation}
where $\mathbf{W}_\theta$ and $\mathbf{W}_\phi$ are learnable parameter of the two embedding functions, which are initialized as $0.0$. Then an AGCN block is realized with a $1\times\Gamma$ classical 2D convolutional layer ($\Gamma$ is the temporal kernel size that we set to $9$) and the defined adaptive graph convolution $\mathbf{f}_{out}(\upsilon_{t}$), which are both followed by a batch normalization layer and a ReLU layer and a dropout layer in between them. Meanwhile, a residual connection \cite{he2016deep} is added to the AGCN block. 

\noindent \textbf{Reconstruct 3D Pose Sequence}.
Taking the inspiration of recent works \cite{RN014,hu2021conditional,li2022mhformer,li2022exploiting},  the introduced AGCN block is then used to extract the spatiotemporal structural information in the global graph representation, which is supervised by estimating the 3D pose sequence from the corresponding 2D sequence (see Figure \ref{fig:architecture_SAGCN} [\textit{Reconstruct 3D Pose Sequence}]).
Here, each AGCN block has three key parameters: the number of input channels $C_{in}$, the number of output channels $C_{out}$, and the stride $S$ of the temporal convolution, while the other parameters are kept consistent (e.g., the temporal convolution kernel size is three). Given an input $C_{in}$-dim pose representation  $F(C_{in}, T_{in}, N)$, the AGCN block derives the output $C_{out}$-dim pose $F(C_{out}, T_{out}, N)$ via convolution on the pose structure sequence, where $T_{out}$ depends on $N_{in}$ and $S$. To reconstruct the 3D pose sequence, we first use $AGCN(2,96,1)$ to convert the 2D pose sequence $F(2,T,N)$ into a 96D pose representation $F(96,T,N)$. Following the settings of related work, we set $T$ to 243 and $N$ to 17 for the Human3.6M dataset. That is, the input 2D pose sequence of $F(2, 243, 17)$ is converted into a 96D pose sequence of $F(96,243,17)$. Then, we stack iterative layers of $AGCN(96,96,1)$ to construct the deep spatiotemporal structural representation of the 96D pose sequence. The output of the last AGCN block is fed into an $AGCN(96,3,1)$ to estimate the 3D pose sequence based on the 96D joint representation and derive $F(3,243,17)$. Then, we let $\dddot{p}_{t,i} \in \mathbb{R}^3$ be the 3D position of the $i$-th joint at time $t$, and minimize the difference between the estimated 3D pose sequence and the ground truth 3D pose sequence:
\begin{equation}
  \label{eq:loss_global}
  \mathcal{L}_{global}=\frac{1}{T} \frac{1}{N} \sum_{t=1}^T \sum_{i=1}^N\left\|\dddot{p}_{t, i}-p_{t, i}\right\|_2
\end{equation}

\noindent \textbf{Strided Learning Architecture}. Inspired by the TCN-based approaches \cite{RN012,RN013}, we further adapt the strided learning architecture to the AGCN model, using strided convolution to reduce long time sequences and aggregate temporal information near time $t$ for pose estimation. The gray module in Figure \ref{fig:architecture_SAGCN}(\textit{Strided Learning}) illustrates the design of the strided AGCN modules. Each strided AGCN module has two consecutive AGCN blocks, which are surrounded by residual connections \cite{he2016deep}. We perform strided convolutions at the second AGCN block of each strided AGCN module to gradually shrink the feature size at the temporal dimension. The input of the first strided AGCN module is the intermediate output in 3D pose sequence reconstruction, i.e., the extracted $F(96, 243,17)$. After the propagation through the first strided AGAN module, the 96D pose sequence will be shrunken to $F(96,81,17)$. Then, we repetitively perform subsequence AGCN layers until the feature size is shrunken to the size of $96\times1\times17$. In this way, the pattern of the temporal neighbor in the pose sequence will be aggregated for subsequent local 3D pose joint estimation to estimate the 3D pose of the centric timestep.

\subsection{Local 3D Pose Joint Estimation}
Based on the above-mentioned strided AGCN modules, the input 2D pose sequence represented as $F(96,243,17)$ can be transformed into a feature map $F(96,1,17)$. The next step is to estimate the 3D position of joint nodes based on the feature map.

\noindent \textbf{Individually Connected Layers}. Existing TCN- and GCN-based methods \cite{RN012,RN013,RN018,RN014} usually flatten the derived feature maps and use a global skeleton representation consisting of all joint nodes to estimate every single joint, neglecting the matching information between joints and corresponding vectors in feature maps. Unlike existing works, we believe the global knowledge of the temporal and spatial neighborhoods has been aggregated via the proposed global representation. Thus, it is crucial to scope at the spatial information of the corresponding joint node to infer its 3D position. Based on this idea, this paper first proposes an individual connected layer to estimate the 3D position of every single joint based on the corresponding joint node feature $F(96,1,1)$, instead of the pooled representation of all joint nodes $F(96,1,17)$. Mathematically, the individual connected layer can be denoted as:
\begin{equation}
  \label{eq:ind_fc_layer}
  \dot{p}_i^{(unshared)}=v_i \mathbf{W}_i+\mathbf{b}_i
\end{equation}
where the estimated 3D position of joint $i$ is denoted by $\dot{p}_i$ and $v_i$ represents the flattened features of $F(96,1,i)$ joint node $i$. The weight parameters of the individual connected layer is represented by $\mathbf{W}_i$ and $\mathbf{W}_i \in \mathbb{R}^{96 \times 3}$, whose bias parameter is $\mathbf{b}_i$ and $\mathbf{b}_i \in \mathbb{R}^{1 \times 3}$.

Due to the weight $\mathbf{W}_i$ and bias $\mathbf{b}_i$ are not shared between joints, we name the above individually connected layers as unshared individually connected layers. On top of that, we find that individually connected layers in the unshared fashion may ignore the shared rules between joints in 2D to 3D pose lifting, resulting in overfitting joint-specific distribution. Therefore, we further designed shared individually connected layers:
\begin{equation}
  \label{eq:shared_fc_layer}
  \dot{p}_i^{(shared)}=v_i \mathbf{W}_{s} + \mathbf{b}_{s}
\end{equation}
The weight parameters of the shared individual connected layer is represented by $\mathbf{W}_{s}$ and $\mathbf{W}_{s} \in \mathbb{R}^{96 \times 3}$, whose bias parameter is $\mathbf{b}_{s}$ and $\mathbf{b}_{s} \in \mathbb{R}^{1 \times 3}$. Then, the 3D pose estimation of each joint can be formulated as the weighted average of the estimated results from the shared and unshared individually connected layers:
\begin{equation}
  \label{eq:final_fc_layer}
  \bar{p}_i=\lambda\dot{p}_i^{(unshared)} + (1-\lambda)\dot{p}_i^{(shared)}
\end{equation}
Here, $\lambda$ is the parameter that weighs the shared individual connected layer and the unshared individual connected layer. When $\lambda$ is 0.0, the model uses only the shared individual connected layer for estimation, and when $\lambda$ is 1.0, the model uses only the unshared individual connected layer for prediction. Especially, for convenience, the connected layers are implemented via a 1D CNN layer in this paper. Finally, we wish to minimize the difference between the estimated joint pose $\bar{p}_{i}$ and the ground truth joint pose $p_{i}$ via $L_{local}$:
\begin{equation}
  \label{eq:loss_local}
  \mathcal{L}_{local}=\frac{1}{N} \sum_{i=1}^N\left\|\bar{p}_{i}-p_{i}\right\|_2
\end{equation}

During the training process, we optimize $\mathcal{L}_{global}$ and $\mathcal{L}_{local}$ in two stages. In the first stage, we minimize $\mathcal{L}_{global}+\mathcal{L}_{local}$ to optimize the model using globally supervised signal guidance. In the second stage, we minimize $\mathcal{L}_{local}$ to improve the 3D pose estimation performance.

\section{Experiments}

\begin{table*}[h]
  \centering
  \resizebox{0.98\linewidth}{!}{%
  \begin{tabular}{lcccccccccccccccc}
    \toprule
	Method  &  Dir.  &  Disc.  &  Eat.  &  Greet  &  Phone  &  Photo  &  Pose  &  Purch.  &  Sit  &  SitD.  &  Smoke  &  Wait  &  WalkD.  &  Walk  &  WalkT.  &  Avg.  \\
    \midrule
Martinez et al. \cite{RN011} (ICCV’17)   &  51.8  &  56.2  &  58.1  &  59.0  &  69.5  &  78.4  &  55.2  &  58.1  &  74.0  &  94.6  &  62.3  &  59.1  &  65.1  &  49.5  &  52.4  &  62.9  \\
Fang et al. \cite{RN034} (AAAI’18)  &  50.1  &  54.3  &  57.0  &  57.1  &  66.6  &  73.3  &  53.4  &  55.7  &  72.8  &  88.6  &  60.3  &  57.7  &  62.7  &  47.5  &  50.6  &  60.4  \\
Pavlakos et al. \cite{RN035} (CVPR’18)   &  48.5  &  54.4  &  54.4  &  52.0  &  59.4  &  65.3  & 49.9  &  52.9  &  65.8  &  71.1  &  56.6  &  52.9  &  60.9  &  44.7  &  47.8  &  56.2  \\
Lee et al. \cite{RN036} (ECCV’18) †    &  40.2  &  49.2  &  47.8  &  52.6  &  50.1  &  75.0  &  50.2  &  43.0  &  55.8  &  73.9  &  54.1  &  55.6  &  58.2  &  43.3  &  43.3  &  52.8  \\
Zhao et al. \cite{RN018} (CVPR’19)    &  47.3  &  60.7  &  51.4  &  60.5  &  61.1  &  49.9  &  47.3  &  68.1  &  86.2  &  \textbf{55.0}  &  67.8  &  61.0  &  \underline{42.1}  &  60.6  &  45.3  &  57.6  \\
Ci et al. \cite{RN029} (ICCV’19)   &  46.8  &  52.3  &  44.7  &  50.4  &  52.9  &  68.9  &  49.6  &  46.4  &  60.2  &  78.9  &  51.2  &  50.0  &  54.8  &  40.4  &  43.3  &  52.7  \\
Pavllo et al. \cite{RN012} (CVPR’19) †   &  45.2  &  46.7  &  43.3  &  45.6  &  48.1  &  55.1  &  44.6  &  44.3  &  57.3  &  65.8  &  47.1  &  44.0  &  49.0  &  32.8  &  33.9  &  46.8  \\
Cai et al. \cite{RN038} (ICCV’19) †  &  44.6  &  47.4  &  45.6  &  48.8  &  50.8  &  59.0  &  47.2  &  43.9  &  57.9  &  61.9  &  49.7  &  46.6  &  51.3  &  37.1  &  39.4  &  48.8  \\
Pavllo et al. \cite{RN012} (CVPR’19) † &  45.2  &  46.7  &  43.3  &  45.6  &  48.1  &  55.1  &  44.6  &  44.3  &  57.3  &  65.8  &  47.1  &  44.0  &  49.0  &  32.8  &  33.9  &  46.8  \\
Xu et al. \cite{RN015} (CVPR’20) † &   \textbf{37.4}  &   43.5  &  42.7  &  42.7  &  46.6  &  59.7  &  41.3  &  45.1  &  \underline{52.7}  &  60.2  &  45.8  &  \textbf{43.1}  &  47.7  &  33.7  &  37.1  &  45.6  \\
Liu et al. \cite{RN013} (CVPR’20) † &  41.8  &  44.8  &  41.1  &  44.9  &  47.4  &  54.1  &  43.4  &  42.2  &  56.2  &  63.6  &  45.3  &  43.5  &  45.3  &  \underline{31.3}  &  32.2  &  45.1  \\
Zeng et al. \cite{zeng2020srnet} (ECCV’20) †  &  46.6  &  47.1  &  43.9  &  41.6  &  45.8  &  49.6  &  46.5  &  40.0  &  53.4  &  61.1  &  46.1  &  42.6  &  43.1  &  31.5  &  32.6  &  44.8  \\
Xu and Takano \cite{RN039} (CVPR’21)  &  45.2  &  49.9  &  47.5  &  50.9  &  54.9  &  66.1  &  48.5  &  46.3  &  59.7  &  71.5  &  51.4  &  48.6  &  53.9  &  39.9  &  44.1  &  51.9  \\
Zhou et al. \cite{RN040} (PAMI’21) †  & \underline{38.5}  &  45.8  &  \underline{40.3}  &  54.9  &  \textbf{39.5}  &  \textbf{45.9}  &  \textbf{39.2}  &  43.1  &  \textbf{49.2}  &  71.1  &  \textbf{41.0}  &  53.6  &  44.5  &  33.2  &  34.1  &  45.1  \\
Li et al. \cite{li2022mhformer} (CVPR’22) †    &  39.2  &  \underline{43.1}  &  \textbf{40.1}  &  \textbf{40.9}  &  \underline{44.9}  &  51.2  &  \underline{40.6}  &  41.3  &  53.5  &  60.3  &  \underline{43.7}  &  41.1  &  43.8  &  29.8  &  30.6  &  \underline{43.0}  \\
Shan et al. \cite{shan2022p} (ECCV’22) †  &  38.9  &  \textbf{42.7}  &  40.4  &  \underline{41.1}  &  45.6  &  \underline{49.7}  &  40.9  &  \textbf{39.9}  &  55.5  &  \underline{59.4}  &  44.9  &  42.2  &  \textbf{42.7}  &  \textbf{29.4}  &  \textbf{29.4}  &  \textbf{42.8}  \\

\rowcolor{grayrow}
Our GLA-GCN (T=243, CPN) †  &  41.3  &  44.3  &  40.8  &  41.8  &  45.9  &  54.1  &  42.1  &  \underline{41.5}  &  57.8  &  62.9  &  45.0  &  \underline{42.8}  &  45.9  &  \textbf{29.4}  &  \underline{29.9}  &  44.4  \\

\midrule\midrule

Martinez et al. \cite{RN011} (ICCV’17)  &  37.7  &  44.4  &  40.3  &  42.1  &  48.2  &  54.9  &  44.4  &  42.1  &  54.6  &  58.0  &  45.1  &  46.4  &  47.6  &  36.4  &  40.4  &  45.5  \\
Lee et al. \cite{RN036} (ECCV’18) †  &  32.1  &  36.6  &  34.3  &  37.8  &  44.5  &  49.9  &  40.9  &  36.2  &  44.1  &  45.6  &  35.3  &  35.9  &  30.3  &  37.6  &  35.5  &  38.4  \\
Zhao et al. \cite{RN018} (CVPR’19)   &  37.8  &  49.4  &  37.6  &  40.9  &  45.1  &  41.4  &  40.1  &  48.3  &  50.1  &  42.2  &  53.5  &  44.3  &  40.5  &  47.3  &  39.0  &  43.8  \\
Ci et al. \cite{RN029} (ICCV’19)  &  36.3  &  38.8  &  29.7  &  37.8  &  34.6  &  42.5  &  39.8  &  32.5  &  \underline{36.2}  &  \underline{39.5}  &  34.4  &  38.4  &  38.2  &  31.3  &  34.2  &  36.3  \\
Liu et al. \cite{RN013} (CVPR’20) † &  34.5  &  37.1  &  33.6  &  34.2  &  32.9  &  37.1  &  39.6  &  35.8  &  40.7  &  41.4  &  33.0  &  33.8  &  33.0  &  26.6  &  26.9  &  34.7  \\
Xu and Takano \cite{RN039} (CVPR’21)  &  35.8  &  38.1  &  31.0  &  35.3  &  35.8  &  43.2  &  37.3  &  31.7  &  38.4  &  45.5  &  35.4  &  36.7  &  36.8  &  27.9  &  30.7  &  35.8  \\
Zheng et al. \cite{zheng20213d} (ICCV’21) †   &  30.0  &  33.6  &  29.9  &  31.0  &  30.2  &  33.3  &  34.8  &  31.4  &  37.8  &  38.6  &  31.7  &  31.5  &  29.0  &  23.3  &  23.1  &  31.3  \\
Li et al. \cite{li2022mhformer}  (CVPR’22) †  &  \underline{27.7}  &  32.1  &  \underline{29.1}  &  28.9  &  30.0  &  33.9  &  33.0  &  31.2  &  37.0  &  39.3  &  30.0  &  31.0  &  29.4  &  22.2  &  23.0  &  30.5  \\
Shan et al. \cite{shan2022p} (ECCV’22) † &  28.5  &  \underline{30.1}  &  \textbf{28.6}  &  \underline{27.9}  &  \underline{29.8}  &  \underline{33.2}  &  \underline{31.3}  &  \underline{27.8}  &  \textbf{36.0}  &  \textbf{37.4}  &  \textbf{29.7}  &  \underline{29.5}  &  \underline{28.1}  &  \underline{21.0}  &  \underline{21.0}  &  \underline{29.3}  \\
\rowcolor{grayrow}
Our GLA-GCN (T=243, GT) † &  \textbf{26.5}  &  \textbf{27.2}  &  29.2  &  \textbf{25.4}  &  \textbf{28.2}  &  \textbf{31.7}  &  \textbf{29.5}  &  \textbf{26.9}  &  37.8  &  39.9  &  \underline{29.9}  &  \textbf{27.0}  &  \textbf{27.3}  &  \textbf{20.5}  &  \textbf{20.8}  &  \textbf{28.5}  \\

\midrule

Wang et al. \cite{RN014} (ECCV’20) †*   &  23.0  &  25.7  &  22.8  &  22.6  &  24.1  &  30.6  &  24.9  &  24.5  &  31.1  &  35.0  &  25.6  &  24.3  &  25.1  &  19.8  &  18.4  &  25.6  \\
Li et al. \cite{li2022exploiting} (TMM’22) †*  &  27.1  &  29.4  &  26.5  &  27.1  &  28.6  &  33.0  &  30.7  &  26.8  &  38.2  &  34.7  &  29.1  &  29.8  &  26.8  &  19.1  &  19.8  &  28.5  \\
Hu et al. \cite{hu2021conditional} (MM’22) †*   &  -  &  -  &  -  &  -  &  -  &  -  &  -  &  -  &  -  &  -  &  -  &  -  &  -  &  -  &  -  &  22.7  \\ 
Zhang et al. \cite{zhang2022mixste} (CVPR’22) †*   &  \underline{21.6}  &  \underline{22.0}  &  \underline{20.4}  &  \underline{21.0}  &  \textbf{20.8}  &  \textbf{24.3}  &  \underline{24.7}  &  \underline{21.9}  &  \textbf{26.9}  &  \textbf{24.9}  &  \underline{21.2}  &  \underline{21.5}  &  \underline{20.8}  &  \underline{14.7}  &  \underline{15.7}  &  \underline{21.6}  \\
\rowcolor{grayrow}
Our method (T=243, GT) † * &  \textbf{20.1}  &  \textbf{21.2}  &  \textbf{20.0}  &  \textbf{19.6}  &  \underline{21.5}  &  \underline{26.7}  &  \textbf{23.3}  &  \textbf{19.8}  &  \underline{27.0}  &  \underline{29.4}  &  \textbf{20.8}  &  \textbf{20.1}  &  \textbf{19.2}  &  \textbf{12.8}  &  \textbf{13.8}  &  \textbf{21.0}  \\
    \bottomrule
  \end{tabular}}
   \caption{$Protocol\#1$: Reconstruction error with MPJPE (mm) on Human3.6M. Top-table: input 2D pose sequences are detected by (CPN) - cascaded pyramid network. Bottom-table: input 2D pose sequences with ground truth (GT). Best in bold, second best underlined, the lower the better. † indicates using temporal information.  * indicates reconstructing an intermediate 3D pose sequence.}
  \label{compare2sota_p1}  
  \vspace{-5pt}
\end{table*}

\begin{table}[ht]
  \centering
    \resizebox{0.98\linewidth}{!}{%
  \begin{tabular}{lccccccc}
    \toprule
   \multirow{2}[3]{*}{Method} & \multicolumn{3}{c}{Walk} & \multicolumn{3}{c}{Jog} &\multirow{2}[3]{*}{Avg} \\
    \cmidrule(lr){2-4} \cmidrule(lr){5-7}
    & S1 & S2 & S3  & S1 & S2 & S3  \\
    \midrule
Martinez et al. \cite{RN011} (ICCV’17)  &  19.7  &  17.4  &  46.8  &  26.9  &  18.2  &  18.6  &  24.6  \\ 
Fang et al. \cite{RN034} (AAAI’18)  &  19.4  &  16.8  &  37.4  &  30.4  &  17.6  &  16.3  &  23.0  \\ 
Pavlakos et al. \cite{RN035} (CVPR’18)   &  18.8  &  12.7  &  29.2  &  23.5  &  15.4  &  14.5  &  19.0  \\ 
Lee et al. \cite{RN036} (ECCV’18) †  &  18.6  &  19.9  &  30.5  &  25.7  &  16.8  &  17.7  &  21.5  \\ 
Pavllo et al. \cite{RN012} (CVPR’19) †   &  13.9  &  10.2  &  46.6  &  20.9  &  13.1  &  13.8  &  19.8  \\ 
Liu et al. \cite{RN013} (CVPR’20) † & 13.1  &  \underline{9.8}  &  \underline{26.8}  &  \textbf{16.9}  &  \underline{12.8}  &  \underline{13.3}  &  \underline{15.5}  \\ 
Zheng et al. \cite{zheng20213d} (ICCV’21) † &  14.4  &  10.2  &  46.6  &  22.7  &  13.4  &  13.4  &  20.1  \\ 
Li et al. \cite{li2022exploiting} (TMM’22) †* &  14.0  &  10.0  &  32.8  &  19.5  &  13.6  &  14.2  &  17.4  \\ 
Zhang et al. \cite{zhang2022mixste} (CVPR’22) †*  &   \underline{12.7}  &  10.9  &  \textbf{17.6}  &  22.6  &  15.8  &  17.0  &  16.1 \\ 
\rowcolor{grayrow}
Ours (T=27, MRCNN) †  &  \textbf{12.5}  &  \textbf{9.1}  &  26.9  &  \underline{18.5}  &  \textbf{12.7}  &  \textbf{12.8}  &  \textbf{15.4}  \\ 
\midrule
\midrule
Li et al. \cite{li2022exploiting} (TMM’22) †* &  \underline{9.7}  &  \underline{7.6}  & \underline{15.8}  &  \underline{12.3}  &  \underline{9.4}  &  \underline{11.2}  &  \underline{11.1}  \\ 
\rowcolor{grayrow}
Ours (T=27, GT) †   &   \textbf{8.7}  &   \textbf{6.8}  &   \textbf{11.5}  &   \textbf{10.1}  &   \textbf{8.2}  &   \textbf{9.9}  &   \textbf{9.2}  \\ 
    \bottomrule
  \end{tabular}}
  \caption{Results of $Protocol\#2$ for HumanEva-I. † uses temporal information. Best in bold, second best underlined. * indicates reconstructing an intermediate 3D pose sequence.}
	\label{compare2sota_p2_HumanEva-I}
 \vspace{-5pt}
\end{table}

\begin{table}[ht]
  \centering
    \resizebox{0.9\linewidth}{!}{%
  \begin{tabular}{lccc}
\toprule
\multicolumn{1}{l}{Method} & PCK↑ & AUC↑ & MPJPE↓ \\ \midrule
Mehta et al. \cite{mehta2017monocular} (3DV’17, T=1)  & 75.7 & 39.3   & 117.6 \\
Pavllo et al. \cite{RN012} (CVPR’19, T=81) † &  86.0 &  51.9 & 84.0  \\
Lin et al. \cite{lin2019trajectory} (BMVC’19, T=25) † & 83.6 & 51.4 & 79.8  \\
Wang et al. \cite{RN014} (ECCV’20, T=96) †*  & 86.9 & 62.1 & 68.1  \\
Zheng et al. \cite{zheng20213d} (ICCV’21, T=9) †   & 88.6 & 56.4  & 77.1  \\
Chen et al. \cite{chen2021anatomy} (TCSVT’21, T=81) †  & 87.9 & 54.0  & 78.8  \\
Hu et al. \cite{hu2021conditional} (MM’22, T=96) †*   &  97.9 & 69.5  & 42.5  \\
Shan et al. \cite{shan2022p} (ECCV’22, T=81) †  &  97.9 & 75.8  & 32.2  \\ 
\rowcolor{grayrow}
Our GLA-GCN (T=27) † & \underline{98.19} & \underline{76.53} & \underline{31.36} \\ 
\rowcolor{grayrow}
Our GLA-GCN (T=81) † & \textbf{98.53} & \textbf{79.12} & \textbf{27.76} \\ \bottomrule
\end{tabular}
    }
  \caption{Results of $Protocol\#1$ for MPI-INF-3DHP. † uses temporal information. Best in bold, second best underlined. * indicates reconstructing an intermediate 3D pose sequence.}
	\label{compare2sota_p2_MPI_INF_3DHP}
 \vspace{-5pt}
\end{table}

\begin{table*}[h]
  \centering
  \resizebox{0.98\linewidth}{!}{%
  \begin{tabular}{lcccccccccccccccc}
    \toprule
	Method  &  Dir.  &  Disc.  &  Eat.  &  Greet  &  Phone  &  Photo  &  Pose  &  Purch.  &  Sit  &  SitD.  &  Smoke  &  Wait  &  WalkD.  &  Walk  &  WalkT.  &  Avg.  \\
    \midrule
Martinez et al. \cite{RN011} (ICCV’17)  &  39.5  &  43.2  &  46.4  &  47.0  &  51.0  &  56.0  &  41.4  &  40.6  &  56.5  &  69.4  &  49.2  &  45.0  &  49.5  &  38.0  &  43.1  &  47.7  \\
Fang et al. \cite{RN034} (AAAI’18)  &  38.2  &  41.7  &  43.7  &  44.9  &  48.5  &  55.3  &  40.2  &  38.2  &  54.5  &  64.4  &  47.2  &  44.3  &  47.3  &  36.7  &  41.7  &  45.7  \\
Pavlakos et al. \cite{RN035} (CVPR’18)  &  34.7  &  39.8  &  41.8  &  38.6  &  42.5  &  47.5  &  38.0  &  36.6  &  50.7  &  56.8  &  42.6  &  39.6  &  43.9  &  32.1  &  36.5  &  41.8  \\
Lee et al. \cite{RN036} (ECCV’18) †  &  34.9  &  \underline{35.2}  &  43.2  &  42.6  &  46.2  &  55.0  &  37.6  &  38.8  &  50.9  &  67.3  &  48.9  &  35.2  &  \textbf{31.0}  &  50.7  &  34.6  &  43.4  \\
Cai et al. \cite{RN038} (ICCV’19) † &  35.7  &  37.8  &  36.9  &  40.7  &  39.6  &  45.2  &  37.4  &  34.5  &  46.9  &  50.1  &  40.5  &  36.1  &  41.0  &  29.6  &  33.2  &  39.0  \\
Pavllo et al. \cite{RN012} (CVPR’19) † &  34.1  &  36.1  &  34.4  &  37.2  &  36.4  &  42.2  &  34.4  &  33.6  &  45.0  &  52.5  &  37.4  &  33.8  &  37.8  &  25.6  &  27.3  &  36.5  \\
Xu et al. \cite{RN015} (CVPR’20) † &  \textbf{31.0}  &  \textbf{34.8}  &  34.7  &  \underline{34.4}  &  36.2  &  43.9  &  \underline{31.6}  &  33.5  &  \textbf{42.3}  &  \underline{49.0}  &  37.1  &  33.0  &  39.1  &  26.9  &  31.9  &  36.2  \\
Chen et al. \cite{RN005} (ICCV’20) †  &  32.9  &  \underline{35.2}  &  35.6  &  \underline{34.4}  &  36.4  &  42.7  &  \textbf{31.2}  &  \underline{32.5}  &  45.6  &  50.2  &  37.3  & 32.8  &  36.3  &  26.0  &  23.9  &  35.5  \\
Liu et al. \cite{RN013} (CVPR’20) † &  \underline{32.3}  &  \underline{35.2}  &  \underline{33.3}  &  35.8 &  35.9  &  \textbf{41.5}  &  33.2  &  32.7  &  44.6  &  50.9  &  37.0  &  \underline{32.4}  &  37.0  &  25.2  &  27.2  &  35.6  \\
Shan et al. \cite{shan2021improving} (MM’21) †  &  32.5  &  36.2  &  33.2  &  35.3  &  \underline{35.6}  &  \underline{42.1}  &  32.6  &  \textbf{31.9}  &  \underline{42.6}  &  \textbf{47.9}  &  \underline{36.6}  &  \textbf{32.1}  &  \underline{34.8}  &  \underline{24.2}  &  \underline{25.8}  &  \underline{35.0}  \\
Shan et al. \cite{shan2022p} (ECCV’22) † &  31.3  &  35.2  &  32.9  &  33.9  &  35.4  &  39.3  &  32.5  &  31.5  &  44.6  &  48.2  &  36.3  &  32.9  &  34.4  &  23.8  &  23.9  &  34.4  \\
Zhang et al. \cite{zhang2022mixste} (CVPR’22) †*  &  30.8  &  33.1  &  30.3  &  31.8  &  33.1  &  39.1  &  31.1  &  30.5  &  42.5  &  44.5  &  34.0  &  30.8  &  32.7  &  22.1  &  22.9  &  32.6  \\
    \rowcolor{grayrow}
Our GLA-GCN (T=243, CPN) †  &  32.4  &  35.3  &  \textbf{32.6}  &  \textbf{34.2}  &  \textbf{35.0}  &  \underline{42.1}  &  32.1  &  \textbf{31.9}  &  45.5  &  49.5  &  \textbf{36.1}  &  \underline{32.4}  &  35.6  &  \textbf{23.5}  &  \textbf{24.7}  &  \textbf{34.8}  \\

    \midrule\midrule
  Martinez et al.  \cite{RN011} (ICCV’17)  &  -  &  -  &  -  &  -  &  -  &  -  &  -  &  -  &  -  &  -  &  -  &  -  &  -  &  -  &  -  &  37.1  \\
Ci et al. \cite{RN029} (ICCV’19)   &  \underline{24.6}  &  \underline{28.6}  &  \underline{24.0}  &  \underline{27.9}  &  \underline{27.1}  &  \underline{31.0}  &  \underline{28.0}  &  \underline{25.0}  &  \underline{31.2}  &  \underline{35.1}  &  \underline{27.6} &  \underline{28.0}  &  \underline{29.1}  &  \underline{24.3}  &  \underline{26.9}  &  \underline{27.9}  \\
    \rowcolor{grayrow}
Our GLA-GCN (T=243, GT) † &  \textbf{20.2}  &  \textbf{21.9}  &  \textbf{21.7}  &  \textbf{19.9}  &  \textbf{21.6}  &  \textbf{24.7}  &  \textbf{22.5}  &  \textbf{20.8}  &  \textbf{28.6}  &  \textbf{33.1}  &  \textbf{22.7}  &  \textbf{20.6}  &  \textbf{20.3}  &  \textbf{15.9}  &  \textbf{16.2}  &  \textbf{22.0}  \\

  \rowcolor{grayrow}
Our GLA-GCN (T=243, GT) †* &  \textbf{16.6}  &  \textbf{18.1}  &  \textbf{16.2}  &  \textbf{17.0}  &  \textbf{17.0}  &  \textbf{22.2}  &  \textbf{19.0}  &  \textbf{17.1}  &  \textbf{22.4}  &  \textbf{25.9}  &  \textbf{17.5}  &  \textbf{16.4}  &  \textbf{16.3}  &  \textbf{10.8}  &  \textbf{11.6}  &  \textbf{17.6}  \\
        \bottomrule
  \end{tabular}}
   \caption{$Protocol\#2$: Reconstruction error  after rigid alignment with P-MPJPE (mm) on Human3.6M. Top-table: input 2D joints are acquired by detection (CPN) - cascaded pyramid network. Bottom-table: input 2D joints with (GT) - ground truth. † indicates using temporal information. * indicates reconstructing an intermediate 3D pose sequence.  Best in bold, second best underlined.}
  \label{compare2sota_p2}  
\end{table*}

\subsection{Datasets and Evaluation}
Our experiments are based on three public datasets: Human3.6M \cite{RN025}, HumanEva-I \cite{RN033}, and MPI-INF-3DHP \cite{mehta2017monocular}. With respect to Human3.6M, the data of subjects S1, S5, S6, S7, and S8 are applied for training, while that of S9 and S11 are used for testing, which is consistent with the training and validation settings of existing works \cite{RN012,RN013,RN018,RN014}. In terms of HumanEva-I, following \cite{RN011} and \cite{RN013}, data for actions “walk” and “jog” from subjects S1, S2, and S3 are used for training and testing. For MPI-INF-3DHP, we follow the experimental setting of the recent state-of-the-art \cite{shan2022p} for a fair comparison. 

Standard evaluation protocols: Mean Per-Joint Position Error (MPJPE) and Pose-aligned MPJPE (P-MPJPE), respectively known as $Protocol\#1$ and $Protocol\#2$, are used for both datasets. The calculation of MPJPE is based on the mean Euclidean distance between the predicted 3D pose joints aligned to root joints (i.e., pelvis) and the ground truth 3D pose joints collected via motion capture, which follows \cite{zhou2016sparseness,tekin2016direct, RN009}. Comparing with MPJPE, P-MPJPE is also based on the mean  Euclidean distance but has an extra post-processing step with rigid alignments (e.g., scale, rotation, and translation) to the predicted 3D pose. P-MPJPE leads to smaller differences with the ground truth and it follows \cite{RN011, hossain2018exploiting,fang2018learning}. 

\subsection{Implementation Details}
We introduce the implementation detail of our GLA-GCN from three main perspectives: 2D pose detections, model setting, and hyperparameters for the training process. For fair comparison, we follow the 2D pose detections of Human3.6M \cite{RN025} and HumanEva-I \cite{RN033} used in \cite{RN012,RN013}, which are detected by CPN \cite{RN005} and MRCNN \cite{RN004}, respectively. The CPN's 2D pose detection has 17 joints while the MRCNN's 2D pose detection has 15 joints. Besides, we also conduct experiments for the ground truth (GT) 2D pose detections of the two datasets.

Based on the specific structure of 2D pose, we implement the graph convolutional operation filters of AGCN blocks,  e.g., the sizes of $\mathbf{A}_k$, $\mathbf{B}_k$, $\mathbf{C}_k$ are set to $17\times17$, $15\times15$,  and $17\times17$ for Human3.6M, HumanEva-I, and MPI-INF-3DHP, respectively. The designed model has some key parameters that can be adjusted to get better performance. For this part, we conduct ablation expariments with difference numbers of channels and 2D pose frames (i.e., $C_{out}$ and $T$, respectively) on Human3.6M. To verify the proper design of the proposed model regarding the strided design and individual connected layer, we perform further ablation experiments on both datasets.

In terms of the hyperparameters, we respectively set the batch size to 512, 256, and 256 for Human3.6M, HumanEva-I, and MPI-INF-3DHP. 
Being consistent with \cite{RN013}, we adopt the ranger optimizer and train the model with the MPJPE loss for $80$ and $1000$ epochs for Human3.6M and HumanEva-I, respectively, using an initial learning rate of $0.01$. Meanwhile, we set the dropout rate to $0.1$. For both training and testing phases, data augmentation is applied by horizontally flipping the pose data. All experiments are conducted with two GeForce GTX 3090 GPUs.

\begin{figure*}[!ht]
	\begin{center}
        \includegraphics[width=0.86\linewidth]{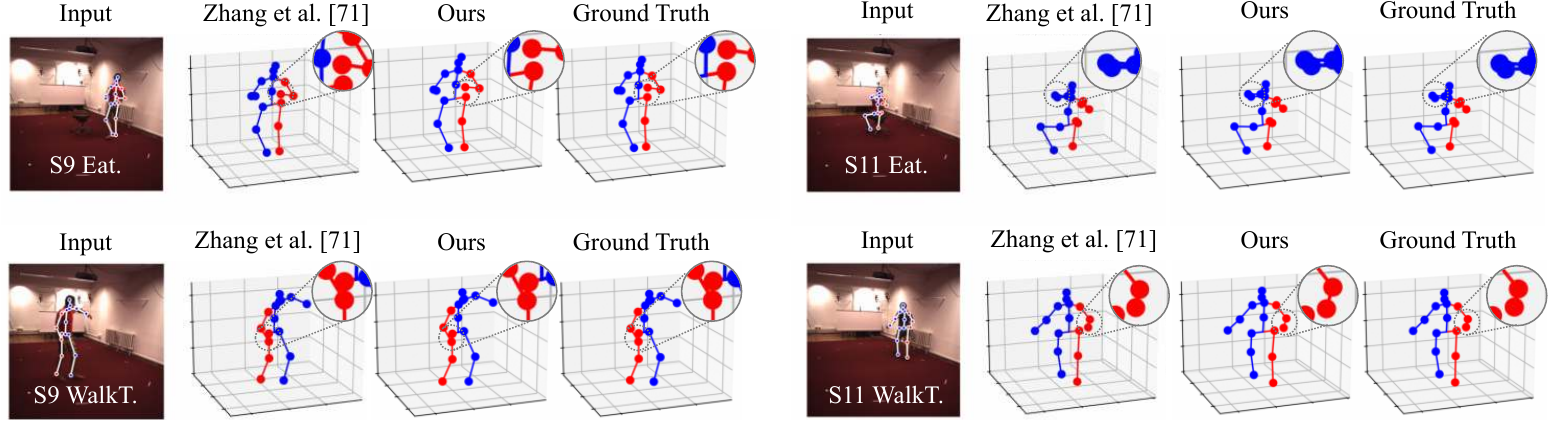}
	\end{center}
  \vspace{-8pt}
	\caption{Qualitative comparison with Zhang et al. \cite{zhang2022mixste} for S9 and S11 on two actions of Human3.6M. Noticeable improvements are enlarged. }
	\label{fig:qualitative_analysis}
	\vspace{-5pt}
\end{figure*}

\begin{figure}[]
  \begin{center}
      \includegraphics[width=\linewidth]{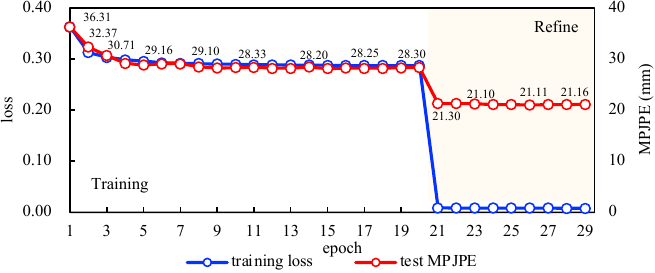}
  \end{center}
  \vspace{-13pt}
  \caption{Loss on the training set and MPJPE on the test set. 
  }
  \vspace{-15pt}
  \label{figure2:supp_train_process}
\end{figure}
\subsection{Comparison with State-of-the-Art}

Tables \ref{compare2sota_p1} and \ref{compare2sota_p2} show the comparison on Human3.6M and HumanEva-I with state-of-the-art methods under $Protocol\#1$ and $Protocol\#2$, respectively. Based on the implementation via GT 2D pose respectively optimized with or without loss of reconstructing the intermediate 3D pose sequence (defined in Equation \ref{eq:loss_global}), our GLA-GCN outperforms the state-of-the-art method \cite{zhang2022mixste} in terms of averaged results of two evaluation protocols. Figure \ref{figure2:supp_train_process} shows the training process of our GLA-GCN on Human3.6M, which indicates our model converges
quickly without observable overfitting.
For the HumanEva-I dataset, the results of $Protocol\#2$ (see Table \ref{compare2sota_p2_HumanEva-I}) also show that our method is superior to state-of-the-art methods by just using the MPJPE loss.  
We also conduct a qualitative comparison with the state-of-the-art method that does not have a 3D pose sequence reconstruction module \cite{RN013}. Figure \ref{fig:qualitative_analysis} shows the visualized improvements over \cite{RN013}. 
For example, in the “S11 WalkT.” action, the visualizations of right-hip and right-hand joints estimated with our method and the ground truth 3D pose are clearly separate but those of \cite{RN013} are connected to each other. Moving on to the MPI-INF-3DHP dataset in Table \ref{compare2sota_p2_MPI_INF_3DHP}, we can see a significant decline in MPJPE with our model. 
Compared with the state-of-the-art method, P-STMO\cite{shan2022p}, the MPJPE of our model decreases from $32.2$mm to $27.76$mm, representing an error reduction of approximately $14\%$.

\begin{figure*}[h]
	\begin{center}
		\includegraphics[width=0.94\linewidth]{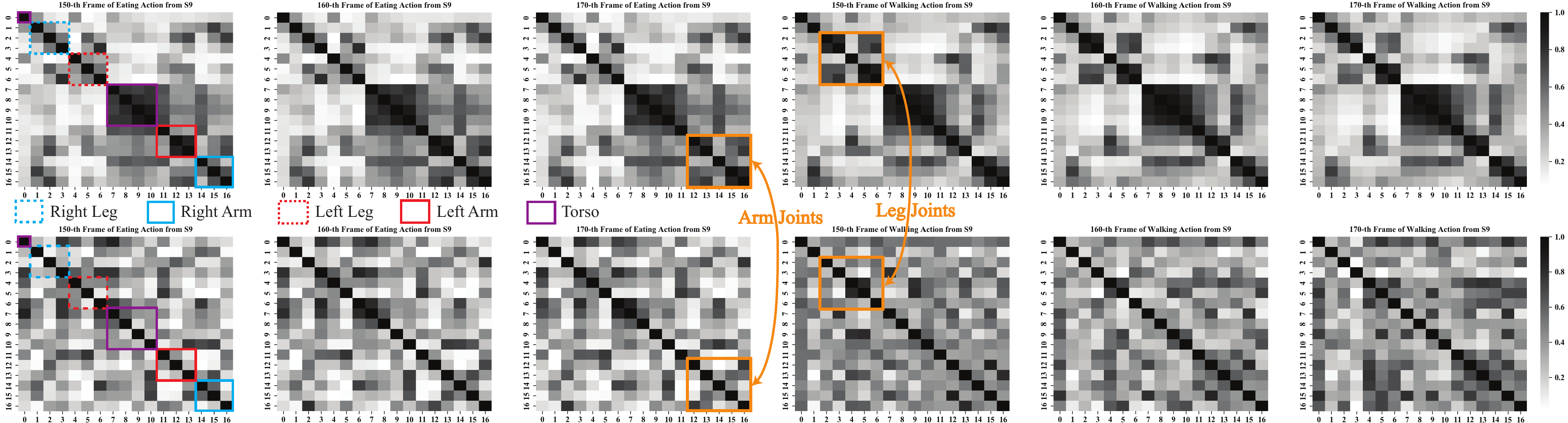}
	\end{center}
	\vspace{-8pt}
	\caption{Visualizations of inter-joint feature cosine similarity from actions: ``eating'' (\textbf{first 3 columns}) and ``walking'' (\textbf{last three columns}) of Human3.6M. \textbf{Upper row} uses an individual connected layer; \textbf{lower row} uses a fully connected layer (please zoom in for a better view). 
	}
	\label{fig:feature_visualization}
 \vspace{-6pt}
\end{figure*}

Comparing the performance by using estimated 2D pose (i.e., CPN or HR-Net pose data) is also regarded as important by most existing works. However, models such as \cite{li2022mhformer,shan2022p} can perform well on relatively low-quality estimated 2D pose data but fail to well generalize the good performance to high-quality 2D data (i.e., ground truth poses). We note that our method underperforms on relatively low-quality estimated 2D pose when compared with some recent methods \cite{li2022mhformer, shan2022p, zhang2022mixste}. In the following, we conduct an in-depth discussion on this issue.

\noindent \textbf{Discussion: the Effect of 2D Pose Quality}.
Tracing back to the first work on 3D Pose lifting, Martinez et al. \cite{RN011} used the SH 2D pose detector fine-tuned on the Human3.6M dataset to improve the 3D HPE (significantly from $67.5$mm average MPJPE to $62.9$mm), indicating that the quality of 2D pose can be essential for the 3D HPE. Recent works \cite{RN014,li2022mhformer,zhang2022mixste} took advantage of advanced 2D pose detector HR-Net and achieved better performance (e.g., $39.8$mm average MPJPE). Zhu et al. \cite{motionbert2022} also successfully improved the result to $37.5$mm average MPJPE by fine-tuning the SH network \cite{RN003} on Human3.6M, which remains far behind the results implemented with GT 2D pose. 

A similar observation is also applicable to the HumanEva-I and MPI-INF-3DHP datasets. As shown in Table \ref{compare2sota_p2_HumanEva-I}, our method yields a remarkable $40\%$ drop in P-MPJPE  on the HumanEva-I dataset. Given the GT 2D pose, the P-MPJPE goes from $15.4$mm to $9.2$mm compared with the best state-of-the-art algorithm. While on  MPI-INF-3DHP, the MPJPE goes from $32.2$mm to $27.76$mm.

Hence, improving the performance of the estimated pose purely relies on preparing quality 2D pose data, which can be easily achieved by either using an advanced 2D pose detector that can generate pose data similar to the GT 2D pose or just arbitrarily fine-tuning the existing pose detectors.
 On the other hand, it remains unclear for what scenario the reconstructed 3D pose with advanced pose detectors can be beneficial. One scenario is 3D human pose estimation in the wild, which is usually evaluated with qualitative visualization \cite{li2022exploiting}. However, whether the 3D pose reconstructed from the estimated 2D pose can contribute to pose-based tasks remains under-explored. 
 Given that how to improve the performance of the estimated 2D pose is straightforward and its usage remains lack of good applicable scenario, we argue that the comparison based on the GT 2D pose can  more properly reflect a model's 3D HPE ability than comparisons based on the estimated 2D pose. 

\begin{table}[ht]
  \centering
    \resizebox{0.98\linewidth}{!}{%
  \begin{tabular}{lccc}
    \toprule
Medhod  &  Frames  &  No. of Parameters  &  MPJPE (mm)  \\
\midrule 
Pavllo et al. \cite{RN012} (CVPR’19) † &  $T$ = 27 &  8.56M  &  40.6  \\
Liu et al. \cite{RN013} (CVPR’20) † &   $T$ = 27  &  \underline{5.69M}  &  38.9  \\
Li et al. \cite{li2022mhformer} (CVPR’22) †* &   $T$ = 27  &  18.92M  &  \textbf{34.3}  \\
  \rowcolor{grayrow}
Our GLA-GCN †  &   $T$ = 27  &  \textbf{0.84M}  &  \underline{34.4}  \\ 
\midrule 
Pavllo et al. \cite{RN012} (CVPR’19) †  &   $T$ = 81  &  12.75M  &  38.7  \\
Liu et al. \cite{RN013} (CVPR’20) † &   $T$ = 81  &  \underline{8.46M}  &  36.2  \\
Li et al. \cite{li2022mhformer} (CVPR’22) †* &   $T$ = 81  &  $\ge 18.92$M  &  \underline{32.7}  \\
  \rowcolor{grayrow}
Our GLA-GCN † &   $T$ = 81  &  \textbf{1.10M}  &  \textbf{31.5}  \\ 
\midrule 
Pavllo et al. \cite{RN012} (CVPR’19) † &   $T$ = 243  &  16.95M  &  37.8  \\
Liu et al. \cite{RN013} (CVPR’20) † &   $T$ = 243  &  \underline{11.25M}  &  \underline{34.7}  \\
\rowcolor{grayrow}
Our GLA-GCN † &   $T$ = 243  &  \textbf{1.35M}  &  \textbf{28.5} 
\\   
\midrule \midrule 
Wang et al. \cite{RN014} (ECCV’20) †*  &   $T$ = 96  &  \textbf{1.69M}  &  25.6  \\   
  \rowcolor{grayrow}
Our GLA-GCN  ($C_{out=96}$) †*  &   $T$ = 243  &  \underline{1.88M}  &  24.5  \\   
Hu et al. \cite{hu2021conditional} (MM’22) †* &   $T$ = 96  &  3.42M  &  22.7  \\   
Li et al. \cite{li2022mhformer} (CVPR’22) †* &   $T$ = 351  &  $\ge 18.92$M  &  30.5  \\
Zhang et al. \cite{zhang2022mixste} (CVPR’22) †*   &   $T$ = 243  &  33.70M  &  \underline{21.6}  \\ 
  \rowcolor{grayrow}
Our GLA-GCN  ($C_{out=512}$) †* &   $T$ = 243  &  48.64M  &  \textbf{21.0}  \\  
\bottomrule
  \end{tabular}}
  \caption{Comparison with state-of-the-art methods on Human3.6M implemented with different receptive fields of ground truth 2D pose. Results of $Protocol\#1$ are reported. * indicates reconstructing an intermediate 3D pose sequence.}
	\label{ablationstudy_T}
 \vspace{-5pt}
\end{table}

\subsection{Ablation Studies}
In the following, we ablate our model design gradients (i.e., AGCN layers, strided design, and individual connected layer).
To validate the properness of using AGCN layers, we compare our model with the version implemented with ST-GCN \cite{RN026} blocks, which leads to the ablation of AGCN. As shown in the row \#1 of Table \ref{tab:ablationstudy_key_design}, the results of $Protocol\#2$ on two datasets consistently indicate that using AGCN blocks can achieve better performances. 

For the ablation of strided design, we perform average pooling on the second (i.e., temporal) dimension of the feature map. Results in row \#2 of Table \ref{tab:ablationstudy_key_design} indicates that it is not as effective as the strided design. Without the strided design, it will not only lead to a larger feature map representation, i.e., increased from $F(C_{out},1,N)$ to $F(C_{out},T,N)$ but also affects the 3D HPE.

To verify the design of our individual connected layer, we compare it with the implementation of a fully connected layer that takes the expanded feature map as its input. The results in the row \#3 of Table \ref{tab:ablationstudy_key_design} indicates that our individual connected layer can make better use of the structured representation of GCN and thus significantly improves the performance. 
The differences of features before the prediction layers (i.e., individually and fully connected layers) are respectively visualized in the upper and lower rows of Figure \ref{fig:feature_visualization}. The visualization indicates that our individual connected layer can make prediction based on more interpretable features that cannot be traced by using a fully connected layer. For example, the arm and leg joints show relatively higher 
independence to other joints for actions ``eating'' and ``walking'', respectively. Feeding these independent features of all joints to a fully connected layer will interfere with the prediction of a specific joint.

We further verify the advantage of this structured representation of GCN by swapping the left and right limbs of 2D pose input data, leading to the break of pose structure. Results in row \#4 of Table \ref{tab:ablationstudy_key_design} show that breaking the pose structure will affect the 3D pose estimation. This observation, in turn, further indicates the proper design of our individual connected layer. 

\begin{table}[t]
  \centering
    \resizebox{0.98\linewidth}{!}{%
  \begin{tabular}{llcccc}
    \toprule
  \multirow{2}[3]{*}{\#} & \multirow{2}[3]{*}{Method} & \multicolumn{2}{c}{Human3.6M} & \multicolumn{2}{c}{HumanEva-I} \\
    \cmidrule(lr){3-4} \cmidrule(lr){5-6}
  &  & CPN & GT   & MRCNN & GT  \\
    \midrule
1 & w/o AGCN  &  39.1  &  28.0  &  18.2  &  \underline{11.7}  \\ 
2 & w/o strided design  &  41.2  &  30.6  &  22.6  & 12.9    \\ 
3 & w/o individual connected layer   &  39.0  &  \underline{27.7}  &  17.6  &  12.4   \\
4 & Swap the structure of input 2D pose &  \underline{38.3}  &  28.1  &  \underline{16.4}  &  12.4  \\ 
\rowcolor{grayrow}
5 & GLA-GCN (T=27) †  &  \textbf{37.8}  &  \textbf{25.8}  &  \textbf{15.4}  &  \textbf{9.2}  \\ 
    \bottomrule
  \end{tabular}}
  \caption{Ablation study on key designs of our GLA-GCN. The results are based on the average value of $Protocol\#2$ implemented with  $27$ receptive fields for various 2D pose detections of the Human3.6M and HumanEva-I datasets.}
	\label{tab:ablationstudy_key_design}
	\vspace{-5pt}
\end{table}

\noindent \textbf{Discussion: Limitation on Model Size}.
Similar to state-of-the-art methods \cite{hu2021conditional, li2022mhformer,zhang2022mixste}, we note that our method is faced with the issue of model size. Specifically, the lower table of Table \ref{ablationstudy_T} shows that our model can achieve better performance than state-of-the-art methods \cite{RN014, zhang2022mixste}  but uses slightly more model parameters. We aim to tackle this issue in the future by using techniques such as pruning.

\section{Conclusion}
This paper proposes a GCN-based method utilizing the structured representation for 3D HPE in the 2D to 3D lifting paradigm. The proposed GLA-GCN globally represents the 2D pose sequence and locally estimates the 3D pose joints via an individual connected layer. Results show that our GLA-GCN outperforms corresponding state-of-the-art methods implemented with GT 2D poses on datasets Human3.6M, HumanEva-I, and MPI-INF-3DHP. We verify the properness of model design with extensive ablation studies and visualizations. In the future, we will tackle the issue of parameter efficiency of our model via tuning techniques \cite{yu2022towards,yu2023visual}. Meanwhile, we will consider its effect on application scenarios such as human behavior understanding \cite{bruce2022mmnet, bruce2021skeleton, bruce2021multimodal,yan2022egcn, bruce2020skeleton} and aim to improve the results of the estimated 2D pose by preparing high-quality 2D pose data via fine-tuned 2D pose detectors (e.g., SH detector \cite{RN003}, OpenPose \cite{RN006}, and  HR-Net \cite{sun2019deep}), abd investigate the effects of other loss terms (e.g., based on bone features \cite{RN019} and motion trajectory \cite{RN014}).

{\small
\bibliographystyle{ieee_fullname}
\bibliography{egbib}
}

\end{document}